\documentclass[conference]{IEEEtran}
\IEEEoverridecommandlockouts
\usepackage{cite}
\usepackage{amsmath,amssymb,amsfonts}
\usepackage{algorithmic}
\usepackage{graphicx}
\usepackage{epsfig}
\usepackage{subfigure}
\usepackage{textcomp}
\usepackage{xcolor}
\usepackage{caption}
\usepackage{diagbox}
\usepackage{multirow}
\usepackage{amsthm}
\usepackage[utf8]{inputenc}
\usepackage{algorithmic}
\usepackage{algorithm}
\usepackage{romannum}
\usepackage{threeparttable}
\usepackage{color,soul}
\def\BibTeX{{\rm B\kern-.05em{\sc i\kern-.025em b}\kern-.08em
    T\kern-.1667em\lower.7ex\hbox{E}\kern-.125emX}}

\begin{document}

\captionsetup[figure]{font={footnotesize},labelformat={default},singlelinecheck={false}, labelsep=period,name={Fig.}}
\captionsetup[table]{justification=centerlast,
                     labelsep=newline,
                     font=small}

\title{Real-time Locational Marginal Price Forecasting Using Generative Adversarial Network}

\author{\IEEEauthorblockN{Zhongxia Zhang}
\IEEEauthorblockA{School of Electrical, Computer and Energy Engineering \\
Arizona State University\\
Tempe, AZ, USA\\
Email: zzhan300@asu.edu}
\and
\IEEEauthorblockN{Meng Wu}
\IEEEauthorblockA{School of Electrical, Computer and Energy Engineering \\
Arizona State University\\
Tempe, AZ, USA\\
Email: mwu@asu.edu}
}

\IEEEoverridecommandlockouts
\IEEEpubid{\makebox[\columnwidth]{978-1-7281-6127-3/20/\$31.00~
\copyright2020
IEEE \hfill} \hspace{\columnsep}\makebox[\columnwidth]{ }}

\maketitle

\begin{abstract}
In this paper, we propose a model-free unsupervised learning approach to forecast real-time locational marginal prices (RTLMPs) in wholesale electricity markets. By organizing system-wide hourly RTLMP data into a 3-dimensional (3D) tensor consisting of a series of time-indexed matrices, we formulate the RTLMP forecasting problem as a problem of generating the next matrix with forecasted RTLMPs given the historical RTLMP tensor, and propose a generative adversarial network (GAN) model to forecast RTLMPs. The proposed formulation preserves the spatio-temporal correlations among system-wide RTLMPs in the format of historical RTLMP tensor. The proposed GAN model learns the spatio-temporal correlations using the historical RTLMP tensors and generate RTLMPs that are statistically similar and temporally coherent to the historical RTLMP tensor. The proposed approach forecasts system-wide RTLMPs using only publicly available historical price data, without involving confidential information of system model, such as system parameters, topology, or operating conditions. The effectiveness of the proposed approach is verified through case studies using historical RTLMP data in Southwest Power Pool (SPP).
\end{abstract}


\section{Introduction}
The forecasting accuracy of wholesale locational marginal prices (LMPs) is critical for electricity market participants to determine optimal trading strategies. As a growing number of renewable generations, energy storage systems, and price-responsive loads are integrated into bulk power systems, the wholesale LMPs become less predictable. Compared to day-ahead LMPs (DALMPs) obtained from a forward market with less price volatility, real-time LMPs (RTLMPs) from a spot market experience more price fluctuation \cite{rule1}, making the RTLMP forecasting a more challenging problem.

To solve the above LMP forecasting problem, various model-based and data-driven methods have been proposed. In \cite{1294980,773811}, LMPs are forecasted using simulation-based approaches which require system generation and transmission models. In \cite{7070121,deng2016probabilistic}, multiparametric programming approaches are applied to forecast LMPs, which assume perfect knowledge of network topology, parameters, and operating conditions. These model-based approaches may be rendered ineffective when applied to the LMP forecasting problems from the market participants' perspective, since market participants could not gain access to confidential system models and operating details. In \cite{7514918,6810300,7478156}, data-driven LMP forecasting methods are developed based on system pattern regions (SPR) \cite{7285827}. These methods forecast LMPs without requiring confidential system model information. The SPRs are highly dependent on system topology, parameters, and generation bids, which are assumed to be constant in \cite{7514918,6810300,7478156}. However, it's not true in the real-world market, since the generation bids vary significantly based on market participants' bidding strategies and major changes may happen in system configurations. Moreover, this SPR-based method takes predicted nodal loads as inputs to forecast future LMP ranges instead of specific LMP values. The LMP forecasting accuracy builds upon the accuracy of predicted nodal loads, which is not guaranteed by real-world markets. In \cite{7917305,8274124,1425583}, time-series statistical models, including the ARMAX model\cite{7917305}, ARIMA model\cite{8274124}, and AGARCH model\cite{1425583}, are applied to forecast LMPs. These time-series models consider linear relationship in historical LMPs and demand uncertainties. However, these methods only model temporal correlations of the historical LMP data for a particular price node (location), without considering spatial correlations among LMPs at various locations (price nodes). Since LMPs across a particular wholesale market are correlated both spatially and temporally, ignoring the spatial correlations among LMPs may negatively affect the LMP forecasting accuracy.

To further improve LMP forecasting accuracy from the market participants' perspective, this paper proposes a model-free unsupervised learning approach to forecast system-wide RTLMPs. The proposed approach doesn't require any system model information. It leverages the spatio-temporal correlations among system-wide historical RTLMPs, and organizes system-wide historical RTLMPs into a three-dimensional (3D) tensor consisting of a series of time-indexed matrices. The RTLMP forecasting problem is then formulated as a problem of generating the next matrix with forecasted RTLMPs, given an input tensor consisting of time-indexed matrices with historical RTLMPs. A generative adversarial network (GAN) model \cite{goodfellow2014generative} is proposed to predict the next matrix for the given tensor, which contains the forecasted RTLMPs at different price nodes (locations) in a wholesale electricity market. The GAN model is trained to learn the nonlinear spatio-temporal correlations among system-wide historical RTLMPs stored in the tensor. A moving-average calibration approach is also proposed to improve the RTLMP forecasting accuracy. Although the proposed approach is applied to forecast real-time LMPs, it can be easily extended to forecasting day-ahead LMPs, as day-ahead LMPs are less volatile \cite{rule1} and therefore more predictable.

The rest of this paper is organized as follows. Section II proposes our formulation of the RTLMP forecasting problem; Section III presents the GAN-based price forecasting approach; Section IV proposes the moving average calibration method for improving forecasting accuracy. Section V verifies the proposed RTLMP forecasting approach through case studies using historical RTLMPs in Southwest Power Pool (SPP) market; Section VI concludes this paper.

\section{Problem Formulation}
In this section, we organize the system-wide historical RTLMPs into a 3D tensor, and formulate the RTLMP forecasting problem as a problem of generating the next matrix given an input series of matrices stored in a tensor. 

\subsection{Representing Historical RTLMPs as A Tensor}
\begin{figure}[h]
    \centering
    \includegraphics[width=9cm]{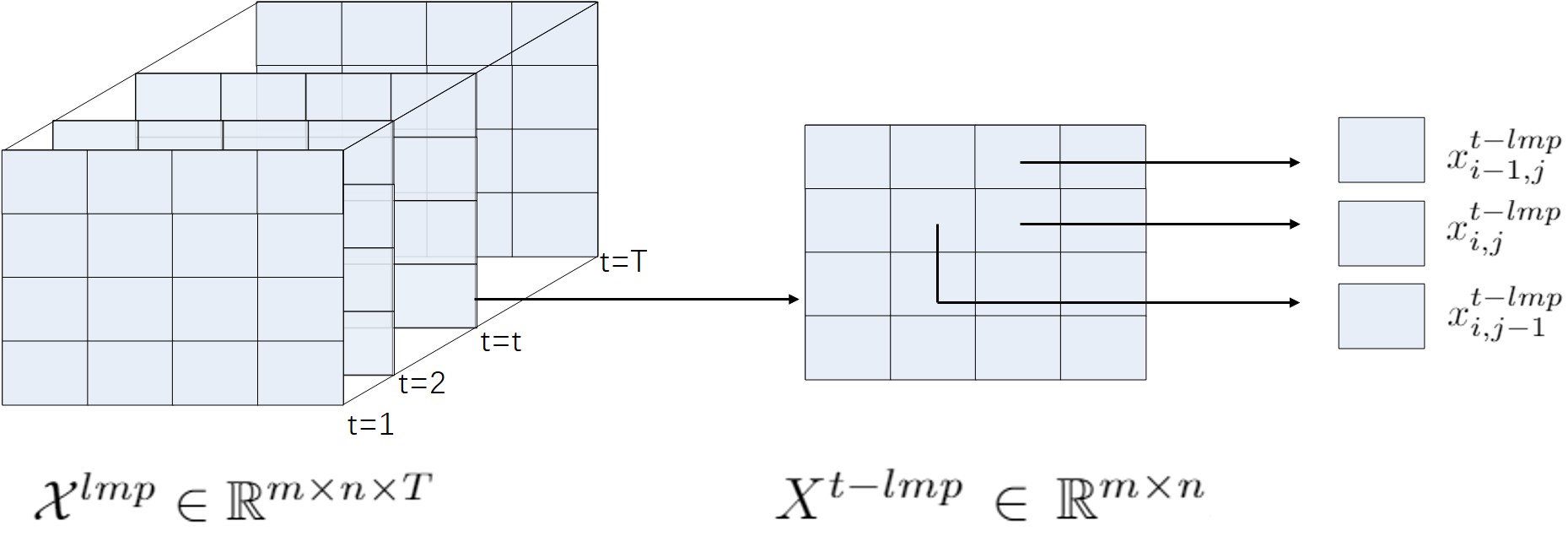}
    \caption{Data structure.}
    \label{fig:datastructure}
\end{figure}

Consider a set of historical hourly RTLMPs collected from $N$ different price nodes (locations) for $T$ consecutive hours. These RTLMP data points can be organized into a tensor $\mathcal{X}^{lmp} \in \mathbb{R}^{m \times n \times T}$, where $m{\times}n=N$ as shown in Fig.~\ref{fig:datastructure}. The tensor $\mathcal{X}^{lmp}$ is comprised of $T$ matrices $X^{t-lmp} \in \mathbb{R}^{m \times n}$, where $t\in [1, T]$. Let $x_{i,j}^{t-lmp}$ be the $(i,j)^{th}$ element of matrix $X^{t-lmp}$. $x_{i,j}^{t-lmp}$ represents the historical RTLMP data collected at time $t$ from $k^{th}$ price node, where $k=n{\times}(i-1)+j$, $k\in[1,N]$.

This tensor representation provides a convenient way to capture inter-dependencies along multiple dimensions. Following the above definitions, the correlations among different elements in the same matrix $X^{t-lmp}$, such as $x_{i,j}^{t-lmp} \in X^{t-lmp}, x_{i-1,j}^{t-lmp} \in X^{t-lmp}, x_{i,j-1}^{t-lmp} \in X^{t-lmp}$, represent the spatial correlations among RTLMPs collected from different price nodes at time $t$; the correlations among $(i,j)^{th}$ elements in different matrices, such as $x_{i,j}^{(t-1)-lmp} \in X^{(t-1)-lmp}, x_{i,j}^{t-lmp} \in X^{t-lmp}, x_{i,j}^{(t+1)-lmp} \in X^{(t+1)-lmp}$, represent the temporal correlations among RTLMPs collected at different hours (from the same price node). The positions of each price node in the 2D array are arranged according to their geographical location in the electricity market footprint.

In above definition, the third dimension of the tensor represents hours. It can be easily modified to represent days. Consider the same set of historical hourly RTLMPs collected from $N$ different price nodes (locations) for $T$ consecutive hours. Let $T=24 \times D$, where $D$ represents total number of days. Within each day, we collect 24 consecutive $X^{t-lmp}$ matrices, and organize them into an enlarged matrix $X^{d-lmp} \in \mathbb{R}^{4m \times 6n}$, where $d \in [1,D]$. The block matrix $X^{d-lmp}$ contains historical hourly RTLMP data collected from $N$ different price nodes within one day (24 hours). The hourly tensor $\mathcal{X}^{lmp}$ is reshaped to a daily tensor $\mathcal{X}_D^{lmp} \in \mathbb{R}^{4m \times 6n \times D}$. Since the daily matrix $X^{d-lmp}$ consists of 24 hourly matrices $X^{t-lmp}$, the correlations among elements in the same daily matrix $X^{d-lmp}$ represent the spatio-temporal correlations among RTLMPs within the same day.

These two types of tensors can be selected flexibly based the forecasting time horizon (hourly and daily). In the following sections, we mainly use hourly tensor structures to present our approaches.

\subsection{Normalizing Historical RTLMPs}
Because the statistics of historical RTLMPs data differ year by year, all historical RTLMPs are preprocessed by normalization. After organizing historical RTLMPs into the tensor $\mathcal{X}^{lmp}$, each element $x_{i,j}^{t-lmp}$ in $\mathcal{X}^{lmp}$ is normalized to the range of -1 and 1, using (\ref{eqn_norm_1})-(\ref{eqn_norm_3}):

\begin{equation}
x_{i,j}^{t-norm}=\frac{ln(x_{i,j}^{t^+})-ln(max(\mathcal{X}^+))/2}{ln(max(\mathcal{X}^+))/2}
\label{eqn_norm_1}
\end{equation}
where
\begin{equation}
x_{i,j}^{t^+}=x_{i,j}^{t-lmp}-min(\mathcal{X}^{lmp})+1
\label{eqn_norm_2}
\end{equation}
\begin{equation}
\mathcal{X}^+=\{X^{1+},\cdots,X^{t+}\,\cdots,X^{T+}\}
\label{eqn_norm_3}
\end{equation}

In (\ref{eqn_norm_1})-(\ref{eqn_norm_3}), $x_{i,j}^{t-norm}$ denotes the normalized value of $x_{i,j}^{t-lmp}$; $x_{i,j}^{t^+}$ denotes the $(i,j)^{th}$ element of matrix $X^{t+}$; $X^{t+}$ denotes the $t^{th}$ matrix in tensor $\mathcal{X}^+$; $max(\mathcal{X}^+)$ denotes the largest element in $\mathcal{X}^+$; $min(\mathcal{X}^{lmp})$ denotes the smallest element in tensor $\mathcal{X}^{lmp}$. 

After normalization, each normalized RTLMP data point, $x_{i,j}^{t-norm}$, lies in the range of -1 to 1. The set of normalized RTLMPs can be represented in tensor $\mathcal{X}^{norm}=\{X^{1-norm},\cdots,X^{t-norm}\,\cdots,X^{T-norm}\}$, where $x_{i,j}^{t-norm}$ denotes the $(i,j)^{th}$ element in matrix $X^{t-norm}$. Equations (\ref{eqn_norm_1})-(\ref{eqn_norm_3}) define a one-to-one mapping between $\mathcal{X}^{lmp}$ and $\mathcal{X}^{norm}$. 

\subsection{Formulating The RTLMP Forecasting Problem}
Following the above data representation, the tensor, $\mathcal{X}^{norm}=\{X^{1-norm},\cdots,X^{t-norm}\,\cdots,X^{T-norm}\}$, contains all the normalized historical RTLMPs at different price nodes (locations) obtained during time $t=1,2,\cdots,T$. We then formulate the RTLMP forecasting problem as the problem of generating a new matrix $X^{(T+1)-norm}$ for the future hour $(T+1)$, such that this newly-generated matrix $X^{(T+1)-norm}$ follows the spatio-temporal correlations in the historical tensor $\mathcal{X}^{norm}$.

Our objective is to train a conditional GAN model consisting of optimal generator and discriminator models, using normalized historical RTLMPs $\mathcal{X}^{norm}=\{X^{1-norm},\cdots,X^{t-norm}\,\cdots,X^{T-norm}\}$ as the training dataset. Throughout the training process, the GAN model learns the sptaio-temporal correlations among the normalized historical RTLMPs in an unsupervised way, and obtains an optimal set of neural network parameters for the generator and discriminator models. This optimal generator model is then used to generate the next matrix $X^{(T+1)-norm}$ following the given input tensor in the third dimension. The generated matrix $X^{(T+1)-norm}$ contains normalized RTLMPs at different price nodes (locations) for the future hour $T+1$.

\section{GAN Model for Price Forecasting}
Built upon the above problem formulation, a time-sequence prediction model is proposed based on deep convolutional GAN. The proposed RTLMP forecasting approach is inspired by solving video prediction problems using GAN, since both video and historical RTLMPs share the tensor format as the input data structure for the time-sequence prediction model. Convolutional GAN model is capable of learning the spatio-temporal correlations stored in tensors and generating new tensors following these correlations. More details on the GAN-based video prediction approach can be found in \cite{mathieu2015deep}.

\subsection{Time-Sequence Prediction Model with GAN}
\begin{figure}[h]
	\centering
	\includegraphics[width=9cm]{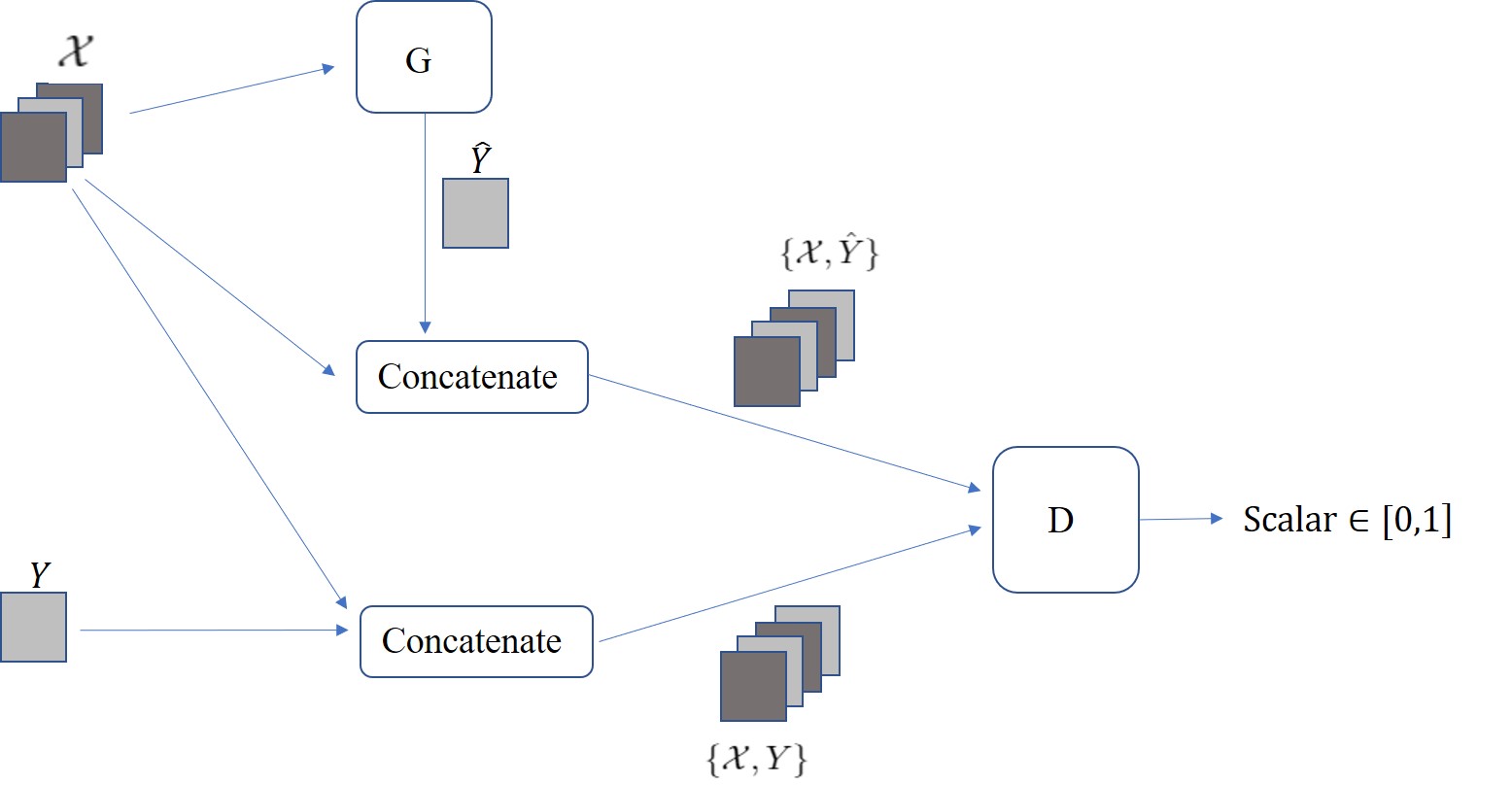}
	\caption{Architecture for training GAN.}
	\label{fig:arch}
\end{figure}

Fig.~\ref{fig:arch} shows the architecture for training the GAN model for RTLMP forecasting. In this architecture, $G$ denotes the generator neural network; $D$ denotes the discriminator neural network; $\mathcal{X}=\{X^1,\cdots,X^n\}$ denotes the tensor consisting series of matrices with normalized historical RTLMPs (at different price nodes/locations) at $n$ consecutive hours, $\mathcal{X}{\subset}\mathcal{X}^{norm}$; Y denotes the matrix with normalized historical RTLMPs at time $n+1$, i.e., $Y=X^{n+1}{\in}\mathcal{X}^{norm}$; $\hat{Y}$ denotes the generated matrix with forecasted normalized RTLMPs for time $n+1$, i.e., $\hat{Y}$ is the forecast of $Y$. 

The generator $G$ takes historical RTLMP tensor $\mathcal{X}=\{X^1,\cdots,X^n\}$ as inputs. It is trained to generate the next-hour RTLMP matrix $\hat{Y}=G(\mathcal{X})$, such that the new concatenated tensor $\{\mathcal{X},\hat{Y}\}$ is statistically similar to the ground-truth RTLMP tensor $\{\mathcal{X},Y\}$.

The discriminator $D$ takes $\{\mathcal{X},\hat{Y}\}$ or $\{\mathcal{X},Y\}$ as the input. The training objective of $D$ is to classify $\{\mathcal{X},\hat{Y}\}$ as fake and $\{\mathcal{X},Y\}$ as real. The output of the discriminator $D$ is a scalar between 0 and 1, indicating the probability of the input tensor being the ground truth.

The generator model, discriminator model, and adversarial training process are described below.

\subsection{The Discriminator $D$}
The discriminator $D$ is a convolutional neural network. It takes a concatenated tensor, $\{\mathcal{X},Y\}$ or $\{\mathcal{X},\hat{Y}\}=\{\mathcal{X},G(\mathcal{X})\}$, as the input. It is then trained to classify the input $\{\mathcal{X},Y\}$ into class $1$ (i.e., $Y$ is classified as the ground-truth matrix) and the input $\{\mathcal{X},\hat{Y}\}=\{\mathcal{X},G(\mathcal{X})\}$ into class $0$ (i.e., $\hat{Y}=G(\mathcal{X})$ is classified as the generated fake matrix). The discriminator $D$ is trained through minimizing the following distance function (loss function):
\begin{align}
\begin{split}
\mathcal{L}_{adv}^D(\mathcal{X},Y)=&\mathcal{L}_{bce}(D(\{\mathcal{X},Y\}),1)\\
&+\mathcal{L}_{bce}(D(\{\mathcal{X},G(\mathcal{X})\}),0)
\end{split}
\end{align}
where $\mathcal{L}_{bce}$ is the following binary cross-entropy:
\begin{equation}
\mathcal{L}_{bce}(k,s)=-[k log(s)+(1-k)log(1-s)]
\label{eqn_cross_entropy}
\end{equation}
where, $k\in [0,1]$ and $s\in \{0,1\}$. The binary cross-entropy measures the distance between the discriminator output $K=D(\{\mathcal{X},\cdot\})$ and the associated label $S$ ($S_i=1$ and $S_i=0$ for real and generated matrices, respectively). These $S$ labels are automatically determined in the training process by the learning algorithm, instead of given with the training dataset, therefore the proposed approach is unsupervised since the original training dataset is unlabeled.

With the above loss function, the discriminator $D$ forces its output scalar $D(\{\mathcal{X},Y\})$ to 1, and $D(\{\mathcal{X},\hat{Y}\})=D(\{\mathcal{X},G(X)\})$ to 0. Note the difference between tensor $\{\mathcal{X},Y\}$ and tensor $\{\mathcal{X},\hat{Y}\}$ is only the last matrix in the third dimension which represents the temporal correlation of RTLMPs between the last hour and previous hours. In this way, the discriminator $D$ takes advantage of learning the temporal correlations in the historical ground-truth tensors and discriminates the ground-truth matrix $Y$ from the fake matrix $\hat{Y}=G(\mathcal{X})$ generated by the generator $G$, given the input ground-truth tensor $\mathcal{X}=\{X^1,\cdots,X^n\}$.

\subsection{The Generator $G$}
The generator $G$ is a generative convolutional neural network. It takes a historical RTLMP tensor, $\mathcal{X}=\{X^1,\cdots,X^n\}$, as inputs and generate a new matrix $\hat{Y}=G(\mathcal{X})$ for the next hour. Given ground truth of the next-hour matrix $Y=X^{n+1}$ and the generated next-hour matrix $\hat{Y}=G(\mathcal{X})$, the objective of the generator $G$ is to minimize a certain distance function (loss function) between the generated matrix $\hat{Y}=G(\mathcal{X})$ and the ground-truth matrix $Y$. This minimization is achieved through adjusting the neural network parameters of $G$ during the training process. In this paper, the following loss function is adopted from \cite{mathieu2015deep} for training $G$:

\begin{align}
\begin{split}
\mathcal{L}^G(\mathcal{X},Y)=&\lambda_{adv}\mathcal{L}_{adv}^G(\mathcal{X},Y)+\lambda_{\ell_p}\mathcal{L}_p(\mathcal{X},Y)\\
&+\lambda_{gdl}\mathcal{L}_{gdl}(\mathcal{X},Y)
\end{split}
\label{eqn_G_distance_full}
\end{align}
where $\mathcal{L}^G(\mathcal{X},Y)$ denotes the multi-loss function for training $G$; $\mathcal{L}_{adv}^G(\mathcal{X},Y)$, $\mathcal{L}_p(\mathcal{X},Y)$, and $\mathcal{L}_{gdl}(\mathcal{X},Y)$ denote three components for this loss function (explained separately in the following sections); $\lambda_{adv}$, $\lambda_{\ell_p}$ and $\lambda_{gdl}$ denote hyperparameters for adjusting the weights of the three loss components.

\subsubsection{The $p$-norm Loss Function $\mathcal{L}_p(\mathcal{X},Y)$}
In (\ref{eqn_G_distance_full}), the following loss function $\mathcal{L}_p(\mathcal{X},Y)$ is introduced to measure the $p$-norm distance between generated matrix $\hat{Y}=G(\mathcal{X})$ and ground-truth matrix $Y$ for the next hour:
\begin{equation}
\mathcal{L}_p(\mathcal{X},Y)=\ell_p(G(\mathcal{X}),Y)=\left\|G(\mathcal{X})-Y\right\|_p^p
\end{equation}
where $\left\|\cdot\right\|_p$ denotes the entry-wise $p$-norm of a particular matrix. When $p=2$ or $p=1$, the above loss function measures the Euclidean or Manhattan distance between $\hat{Y}$ and $Y$, respectively. Intuitively, this loss function measures the Euclidean or Manhattan distance between the ground-truth RTLMPs for the next hour (stored in $Y$) and the forecasted RTLMPs for the next hour (stored in $\hat{Y}$). In the training process, this loss function forces the generator $G$ to forecast next-hour RTLMPs that are close to their ground-truth values.

\subsubsection{The Adversarial Loss Function $\mathcal{L}_{adv}^G(\mathcal{X},Y)$}
To further improve the forecasting accuracy, the following adversarial loss function $\mathcal{L}_{adv}^G(\mathcal{X},Y)$ is integrated into (\ref{eqn_G_distance_full}).
\begin{align}
\begin{split}
\mathcal{L}_{adv}^G(\mathcal{X},Y)&=\mathcal{L}_{bce}(D(\{\mathcal{X},G(\mathcal{X})\}),1)
\end{split}
\end{align}
Over the training process, this adversarial loss function forces the generator $G$ to generate $\hat{Y}$ that is temporally coherent with its input historical RTLMPs tensor $\mathcal{X}=\{X^1,\cdots,X^n\}$ and realistic enough to confuse the discriminator $D$. Details on the adversarial loss function can be found in \cite{mathieu2015deep}.

\subsubsection{The Gradient Difference Loss Function $\mathcal{L}_{gdl}(\mathcal{X},Y)$}
To further utilize the spatial correlations among historical RTLMPs at different locations (price nodes) stored in matrices, the following gradient difference loss function is adopted to learn the spatial correlations \cite{mathieu2015deep}:
\begin{align}
\begin{split}
\mathcal{L}_{gdl}(\mathcal{X},Y)&=\sum_{i,j}\vert\vert Y_{i,j}-Y_{i-1,j}\vert-\vert\hat{Y}_{i,j}-\hat{Y}_{i-1,j}\vert \vert \\
&+\vert \vert Y_{i,j-1}-Y_{i,j}\vert-\vert\hat{Y}_{i,j-1}-\hat{Y}_{i,j}\vert\vert 
\end{split}
\end{align}
where $Y_{i,j}$ and $\hat{Y}_{i,j}$ denote $(i,j)^{th}$ elements in the ground-truth matrix $Y$ and the generated matrix $\hat{Y}=G(\mathcal{X})$, respectively. This gradient difference loss function considers the differences between neighboring elements in a matrix, which represents the spatial correlation between historical RTLMPs obtained at nearby locations (price nodes). For hourly and daily RTLMP forecasts (with hourly and daily tensor structures), this minimization ensures the generated hourly RTLMP matrix captures the spatial correlations and the generated daily RTLMP matrix captures the spatio-temporal correlations within one day, respectively.

\subsection{Adversarial Training}
The above GAN model (in Fig. \ref{fig:arch}) is trained through the adversarial training procedure. The training data set $\mathcal{X}^{norm}$ consists of year-long historical RTLMPs (after normalization) obtained from different locations (price nodes) in a certain wholesale market. The parameters for generator $G$ and discriminator $D$ are updated iteratively. The stochastic gradient decent (SGD) minimization is adopted for obtaining optimal parameters for $G$ and $D$. Algorithm \ref{alg1} shows the overall adversarial training procedure. In each adversarial training iteration, $G$ or $D$ are updated with $M$ different sets of samples (i.e., $M$ historical RTLMP tensors $\mathcal{X}$) one after another. Upon convergence of the training process, $G$ can produce realistic and accurate RTLMP forecasts which cannot be discriminated by $D$. More details on the adversarial training algorithm and the neural network parameters can be found in \cite{mathieu2015deep,URLDCGAN}.

\begin{algorithm}
\caption{Training GAN for forecasting RTLMPs}
\label{alg1}
\begin{algorithmic}
\REQUIRE set the learning rates $\rho_D$ and $\rho_G$, loss hyperparameters $\lambda_{adv}$, $\lambda_{\ell_p}$, $\lambda_{gdl}$, and size of the training data samples for each iteration $M$
\REQUIRE initial discriminative model weights $W_D$ and generative model weights $W_G$
\WHILE{not converged}
\STATE \textbf{Update the discriminator D:}
\STATE Get $M$ data samples from the training data set $\mathcal{X}^{norm}$:
\STATE $(\mathcal{X},Y)=(\mathcal{X}^{(1)},Y^{(1)}),\cdots,(\mathcal{X}^{(M)},Y^{(M)}) \subset \mathcal{X}^{norm}$
\STATE Do one SGD update step
\STATE $W_D=W_D-\rho_D \sum_{i=1}^M \frac{\partial \mathcal{L}_{adv}^D(\mathcal{X}^{(i)},Y^{(i)})}{\partial W_D}$
\STATE \textbf{Update the generator G:}
\STATE Get \textit{new} $M$ data samples the training data set $\mathcal{X}^{norm}$:
\STATE $(\mathcal{X},Y)=(\mathcal{X}^{(1)},Y^{(1)}),\cdots,(\mathcal{X}^{(M)},Y^{(M)}) \subset \mathcal{X}^{norm}$
\STATE Do one SGD update step
\STATE $W_G=W_G-\rho_G \sum_{i=1}^M( \lambda_{adv}\frac{\partial \mathcal{L}_{adv}^G(\mathcal{X}^{(i)},Y^{(i)})}{\partial W_G}+\lambda_{\ell_p}\frac{\partial \mathcal{L}_{\ell_p}(\mathcal{X}^{(i)},Y^{(i)})}{\partial W_G}+\lambda_{gdl}\frac{\partial \mathcal{L}_{gdl}(\mathcal{X}^{(i)},Y^{(i)})}{\partial W_G})$
\ENDWHILE
\end{algorithmic}
\end{algorithm}

\section{Moving Average Calibration}
To apply the above GAN model for the RTLMP forecasting problem, we train the GAN model using year-long historical RTLMPs and then apply the trained GAN model to forecast RTLMPs hour by hour or day by day for the next year. Due to generation/transmission system upgrades and load growth, the statistical distributions of RTLMPs obtained at different years may deviate from each other. To compensate these deviations and further improve the forecasting accuracy, we propose the following moving average calibration for the outputs of the generator neural network $G$:

\begin{equation}
\widetilde{Y}^{t+1}=\widehat{Y}^{t+1}-\frac{\sum_{i=t-3}^{t} (\widehat{Y}^i-Y^i)}{4}
\label{eqn_calibaration}
\end{equation}
where $Y^i$ denotes the matrix with ground-truth RTLMPs at hour $i$; $\widetilde{Y}^{t+1}$ denotes the matrix with forecasted RTLMPs at hour ${t+1}$; $\widehat{Y}^{i}$ denotes the matrix generated by the generator $G$ at hour $i$. Using (\ref{eqn_calibaration}), we calibrate the forecasted RTLMPs for the next hour with the average difference between the RTLMPs generated by $G$ and the ground-truth RTLMPs over the past four hours. Similarly in day-ahead forecasting, forecasted RTLMPs for next day are calibrated with average difference between the generated RTLMPs and the ground-truth over the past four days.

\section{Case Study}
The proposed methodology is applied to forecast RTLMPs in SPP \cite{SPPdata}. Nine price nodes in SPP market are selected. Two cases are studied: Case 1 forecasts hourly RTLMPs in a hour-ahead manner (using the hourly tensor structure). Case 2 forecasts hourly RTLMPs in a day-ahead manner (using the daily tensor structure). The hourly RTLMP forecast accuracy is measured by the mean absolute percentage error (MAPE). The test case data and tensor structures are described as follows.

\subsubsection{Case 1}
The training dataset contains hourly SPP RTLMP data of 9 nodes from 6/1/2016 (00:00:00) to 7/30/2017 (23:00:00). These RTLMP data points are organized into training tensor $\mathcal{X}_{train-case 1} \in \mathbb{R}^{3 \times 3 \times 10224}$. The trained generative model is tested by forecasting hourly RTLMPs for the 9 nodes in a hour-ahead manner in the following four time windows: 7/31/2017-8/13/2017, 8/21/2017-9/3/2017, 9/18/2017-10/1/2017, and 10/2/2017-10/15/2017.

\subsubsection{Case 2}
The basic training dataset is the same as that in Case 1. These RTLMP data points are organized into training tensor $\mathcal{X}_{train-RTLMP} \in \mathbb{R}^{12 \times 18 \times 426}$. To demonstrate advantage of our proposed approach and make a fair comparison with the forecasting approaches in \cite{8733097}, public market data used in \cite{8733097} including day-ahead LMP (DALMP), demand and generation mix data of 9 nodes are added as additional inputs. These additional data points are organized into training tensor $\mathcal{X}_{train-DALMP} \in \mathbb{R}^{12 \times 18 \times 426}$, $\mathcal{X}_{train-D} \in \mathbb{R}^{12 \times 18 \times 426}$ and $\mathcal{X}_{train-GM} \in \mathbb{R}^{12 \times 18 \times 426}$. These additional tensors follow the same structure with $\mathcal{X}_{train-RTLMP}$. According to the time index, these four training tensors are merged into the training tensor for Case 2, $\mathcal{X}_{train-case 2} \in \mathbb{R}^{12 \times 18 \times 1704}$. The trained generative model is tested by forecasting hourly RTLMPs for the same 9 nodes in a day-ahead manner using the same four time windows tested in Case 1.


\subsection{Neural Network Architecture and Configurations}

The proposed models are implemented by Tensorflow \cite{tensorflow} and trained on Google Colab using online GPU for acceleration. Inspired by the state of art video prediction model\cite{mathieu2015deep}, both the generator $G$ and discriminator $D$ are modeled using deep convolutional neural networks excluding any pooling/subsampling layers. The neural network architecture details for $G$ and $D$ of both cases are listed in Table.~\ref{network architecture}.

The generator $G$ takes a historical training tensor ($\mathcal{X}_{case 1} \in \mathbb{R}^{3 \times 3 \times 4 }\subset \mathcal{X}_{train-case 1}$ in Case 1 or $\mathcal{X}_{case 2} \in \mathbb{R}^{12 \times 18 \times 16} \subset \mathcal{X}_{train-case 2}$ in Case 2) obtained over 4 consecutive hours (in Case 1) or days (in Case 2) as the inputs, and forecasts the RTLMP matrix for the next hour (in Case 1) or next day (in Case 2). The forecast is performed by generating the RTLMP matrix for the next hour (in Case 1) or next day (in Case 2), using the training tensor for the past 4 hours (in Case 1) or days (in Case 2). All convolution transpose layers (Conv2DTranspose) in $G$ are followed by batch normalization layers and ReLU units, while the convolution and fully connected layers (Conv2D and Dense) in $D$ are followed by batch normalization layers, Leaky-ReLU units and dropout layers.
\begin{table}[!h]
	\centering  
	\caption{Neural Network Architecture Details}  
	\begin{tabular}{>{\centering\arraybackslash}p{0.9cm} | >{\centering\arraybackslash}p{3.3cm} >{\centering\arraybackslash}p{3.3cm}}  
		\hline 
		\hline
		  Case 1 & Generator G & Discriminator D \\
		  & (Layer Type, Feature Map) & (Layer Type, Feature Map)\\  
		\hline  
		Input  & $3\times 3\times 4$ & $3\times 3\times 5$ \\  
		Layer 1 & Conv2DTranspose, 64 & Conv2D, 64 \\  
		Layer 2 & Concatenate, 256 & Concatenate, 320 \\
		Layer 3 & Conv2DTranspose, 1024 &  Dense, 1024 \\
		Layer 4 & Conv2DTranspose, 512 & Dense, 512 \\
		Layer 5 & Conv2DTranspose, 64 & Dense, 256 \\
		Output & $3\times 3\times 1$ & scalar$\in[0,1]$\\
		\hline
		\hline
		  Case 2 & Generator G & Discriminator D \\
		  & (Layer Type, Feature Map) & (Layer Type, Feature Map)\\  
		\hline  
		Input  & $12\times 18\times 16$ & $12\times 18\times 5$ \\  
		Layer 1 & Conv2DTranspose, 64 & Conv2D, 64 \\  
		Layer 2 & Concatenate, 128 & Concatenate, 320 \\
		Layer 3 & Conv2DTranspose, 256 &  Dense, 1024 \\
		Layer 4 & Conv2DTranspose, 128 & Dense, 512 \\
		Layer 5 & Conv2DTranspose, 64 & Dense, 256 \\
		Output & $12\times 18\times 1$ & scalar$\in[0,1]$\\
		\hline
		\hline
	\end{tabular}
	\label{network architecture}
\end{table}  

Convolution transpose layers in $G$ are with kernel size of $3\times 3$, stride size of $1\times 1$ in Case 1, and kernel size of $12\times 18$, stride size of $1\times 1$ in Case 2. All convolution transpose layers are padded. Convolution layers in $D$ use the same kernel and stride sizes as those used in $G$ for both cases, but are not padded. In the D models, the dropout rates are set to 0.3, the small gradients are set to 0.2 when Leaky-ReLU is not active. In our cases, the proposed model is trained using standard stochastic gradient descent (SGD) optimizer. The size of the data samples $M$ (minibatch size) is set to 4 in Algorithm \ref{alg1}. In the model of Case 1, the learning rates $\rho_G$ and $\rho_D$ are 0.0005, without decay and momentum. In the model of Case 2, the learning rates $\rho_G$ and $\rho_D$ are 0.000005 and 0.00001, without decay and momentum. The loss weight hyperparameters in (\ref{eqn_G_distance_full}) are set to $\lambda_{adv}=0.2$, $\lambda_{\ell_p}=1$, and $\lambda_{gdl}=1$. More details on the neural network structures can be found in \cite{URLDCGAN}.

\subsection{Case Study Results}
\begin{figure}[h]
    \centering
    \includegraphics[width=9cm]{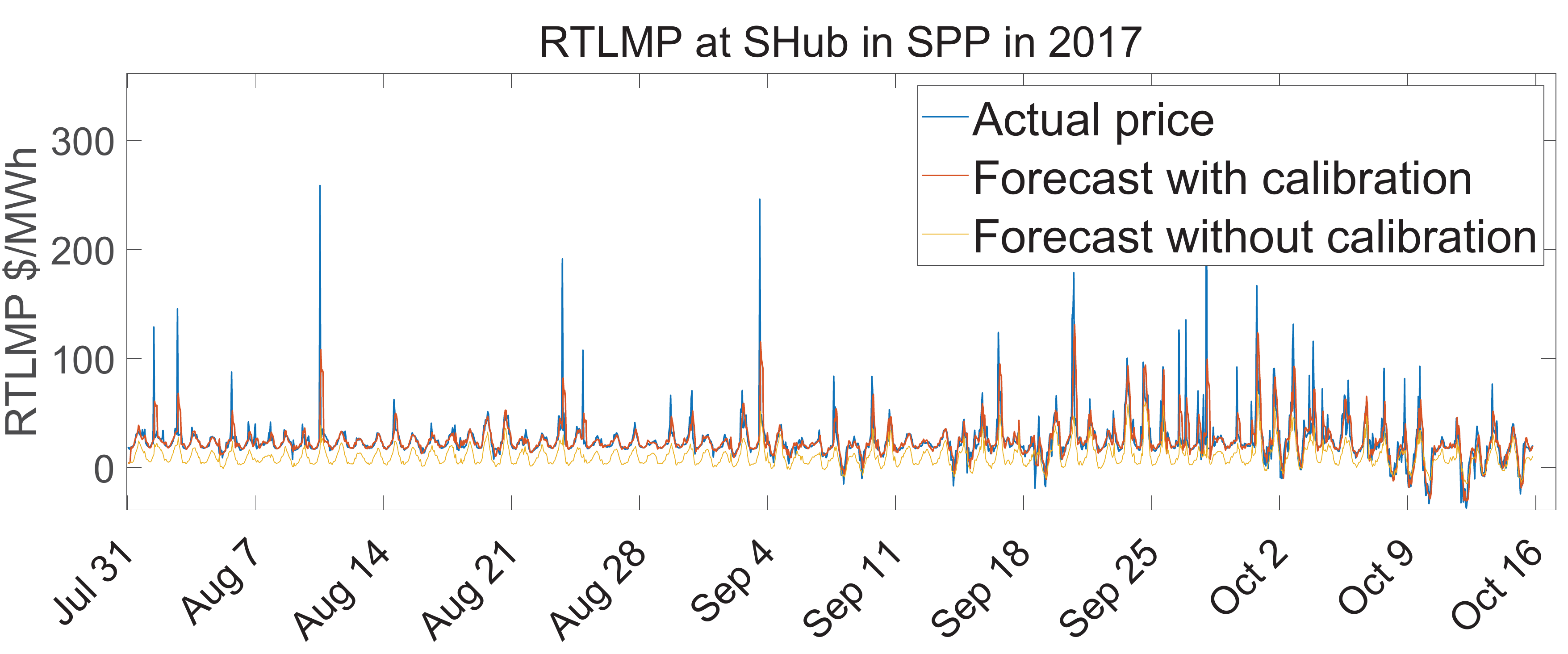}
    \caption{Ground-truth and forecasted RTLMPs (with and without calibration) at South Hub (SHub) price node in SPP.}
    \label{fig:result}
\end{figure}

The trained generator $G$ in Case 1 is employed to forecast hourly RTLMPs in 2017. Fig.~\ref{fig:result} compares the ground-truth RTLMPs and forecasted RTLMPs at South Hub price node in SPP over whole testing period.
 It's clear that without the moving average calibration, the RTLMPs forecasted by the proposed GAN model successfully capture the temporal correlations in the ground-truth RTLMPs. However, there exists a constant bias between the ground-truth RTLMPs and the RTLMPs forecasted without calibration. After applying the moving average calibration, this bias is corrected and the forecasting accuracy is improved. It's clear that the forecasted RTLMPs after calibration closely follow the overall trends of the ground-truth RTLMPs and reflect the correct temporal characteristics.

Table~\ref{accuracy} shows the RTLMP forecast accuracy of the proposed method in Case 1, Case 2, and the MAPEs obtained by two other approaches (ALG+$\hat{M}$ and Genscape) in \cite{8733097} using the same testing data at South Hub and North Hub in SPP real-time market.  ALG+$\hat{M}$, Genscape and Case 2 forecast hourly RTLMPs in a day-ahead manner; Case 1 forecasts hourly RTLMPs in a hour-ahead manner. ALG+$\hat{M}$, Case 1 and Case 2 forecast RTLMPs using only publicly available data; Genscape, which is a commercial product for RTLMP forecasting, incorporates richer and proprietary data which is confidential to market participants. We observe that our proposed approach has a comparable performance to the state-of-art industry benchmark Genscape, using only limited data. Our GAN-based forecast model performs better than the latest existing approach in \cite{8733097}, using the same public market data.
\begin{table}[!h] 
	\centering  
	\begin{threeparttable}
		\caption{RTLMP Prediction Accuracy in Study Cases and \cite{8733097}}  
		\begin{tabular}{>{\centering\arraybackslash}p{1.5cm}|>{\centering\arraybackslash}p{2.5cm} >{\centering\arraybackslash}p{2.5cm}}  
			\hline 
			\hline
			Approach & MAPE (\%) for & MAPE (\%) for \\
				     & SHub Price Zone & NHub Price Zone \\    
			\hline
			ALG+$\hat{M}$\tnote{1} & 25.4 & 36.9\\
			Genscape\tnote{2} & 21.7 & 28.2 \\
			Case 1  &  19.6 &  21.5 \\
			Case 2  &  24.8 &  25.1 \\
			\hline
			\hline
		\end{tabular}
		\begin{tablenotes}
			\footnotesize
			\item[1] The proposed method with the best performance in \cite{8733097}
			\item[2] State of art baseline prediction from Genscape\cite{8733097}
		\end{tablenotes}
		\label{accuracy}
	\end{threeparttable}
\end{table}  



\begin{figure}[h]
	\centering
	\includegraphics[width=8.0cm, height=4cm]{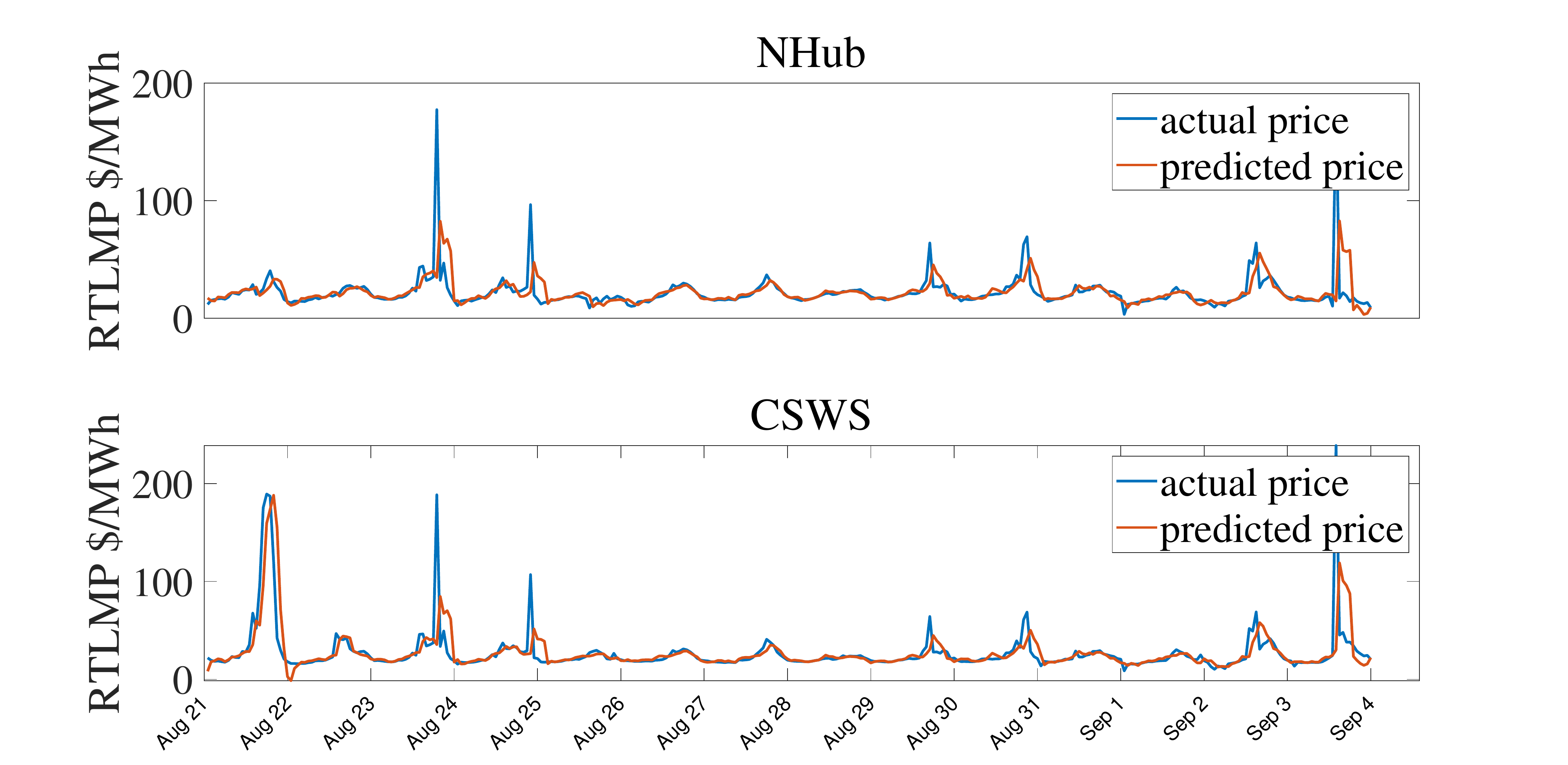} 
	\caption{Ground-truth and Forecasted RTLMPs at North Hub (NHub) and CSWS price nodes in Case 1.}
	\label{sppplot}
\end{figure}

Fig.~\ref{sppplot} shows the ground-truth and forecasted RTLMPs at two different price nodes (the North Hub and CSWS) for the testing window of 8/21/2017-9/3/2017 in Case 1. At both price nodes, the forecasted RTLMPs closely follow the corresponding ground-truth RTLMPs. The proposed approach successfully captures the spatio-temporal correlations of RTLMPs at different price nodes across SPP. 

\section{Conclusion and Future Work}
This paper proposes a GAN-based approach to forecast system-wide RTLMPs. Taking advantage of the deep convolutional neural network and the adversarial training procedure, this approach successfully captures the spatio-temporal correlations among historical RTLMPs. Case studies on system-wide RTLMPs in SPP demonstrate the forecasting accuracy of the proposed approach. Generalization performance of the proposed GAN-based forecasting approach is sensitive to the choice of hyper-parameters and the stopping criterion, which requires further studies. In the future studies, the proposed approach will be compared with other representative methods.

\bibliographystyle{IEEEtran}
\bibliography{reference}

\end{document}